\documentclass[11pt]{article}

\usepackage[final]{acl}

\usepackage{times}
\usepackage{latexsym}

\usepackage{booktabs}
\usepackage[T1]{fontenc}
\usepackage[utf8]{inputenc}
\usepackage{microtype}
\usepackage{inconsolata}
\usepackage{graphicx}
\usepackage[most]{tcolorbox}
\usepackage{xcolor}
\usepackage{bm}

\usepackage{caption}

\DeclareCaptionType{prompt}[Prompt][List of Prompts]
\DeclareCaptionType{instruction}[Instruction][List of Instructions]
\newtcolorbox{llmprompt}{
  colback=gray!10,
  colframe=blue!50!black,
  fonttitle=\bfseries,
  rounded corners,
  breakable,
}

\title{DiSARM: Diverse and Surprising Association Rule Mining with Large Language Models}

\author{
Lijia Liu, Hisashi Kashima\\
Kyoto University \\
}

\begin{document}
\maketitle
\begin{abstract}
Exploratory Data Analysis (EDA) aims to discover informative patterns in large datasets. For transactional data, Association Rule Mining (ARM) is an established approach that facilitates EDA by efficiently generating candidate patterns. 
Beyond efficient mining, identifying informative rules has also been extensively studied through both objective and subjective interestingness measures.
Recent advances in Large Language Models (LLMs) offer new opportunities for assessing subjective interestingness using world knowledge.
Nevertheless, relying solely on LLMs raises concerns about scalability and statistical reliability. 
Building on prior ARM research, we propose a hybrid framework, DiSARM, that leverages the world knowledge of LLMs to assess the subjective interestingness of rules, while maintaining statistical reliability and promoting diversity among the selected rules.
Across three transactional datasets, experiments demonstrate that DiSARM effectively combines statistical reliability with surprise and diversity, offering a promising approach to ADE from transactional data.
\end{abstract}

\section{Introduction}
With the unprecedented growth of data, there is an increased need for methods to analyze large datasets. Exploratory Data Analysis (EDA) aims to discover informative patterns in vast datasets, helping analysts to identify useful knowledge. However, because EDA requires both knowledge of the dataset and domain expertise, the exploration process is time-consuming and requires human supervision. As a result, an emerging body of work focuses on Automated Data Exploration (ADE), which aims to automate the process of EDA and reduce the workload of human analysts.

For transactional data, Association Rule Mining (ARM) is a widely adopted approach to discover association relationships between items~\cite{agrawa1993}. Classical ARM algorithms, such as Apriori~\cite{apriori} and FP-Growth~\cite{fpgrowth}, utilize user-provided support and confidence thresholds to efficiently search large transactional databases and discover associations as clearly interpretable rules. 
ARM has been applied to various fields, including recommendation engines~\cite{rec_liu2025, maulana2025}, customer behavior~\cite{hunyadi2025, siam2025}, enterprise architecture~\cite{EA2024, pinheiro2021}, and healthcare~\cite{jhanjhi2024, hu2022}.
Its ability to efficiently discover interpretable patterns makes ARM a promising approach for generating candidate patterns that can support ADE.

Beyond efficient mining, identifying interesting associations among the discovered rules has also been extensively studied, and research has shown that rules can be evaluated using both objective and subjective interestingness measures~\cite{eldin2026}. Objective interestingness measures quantify statistical strengths of associations in a dataset, such as reliability and generality~\cite{tan2004, hamalainen2008}. Subjective interestingness measures, on the other hand, relate to the benefit acquired by using the discovered pattern, as well as surprisingness and novelty~\cite{liu2000, shekar2004}.

\begin{figure*}[t]
    \centering
    \includegraphics[width=\textwidth, trim={7mm 0mm 7mm 0mm}]{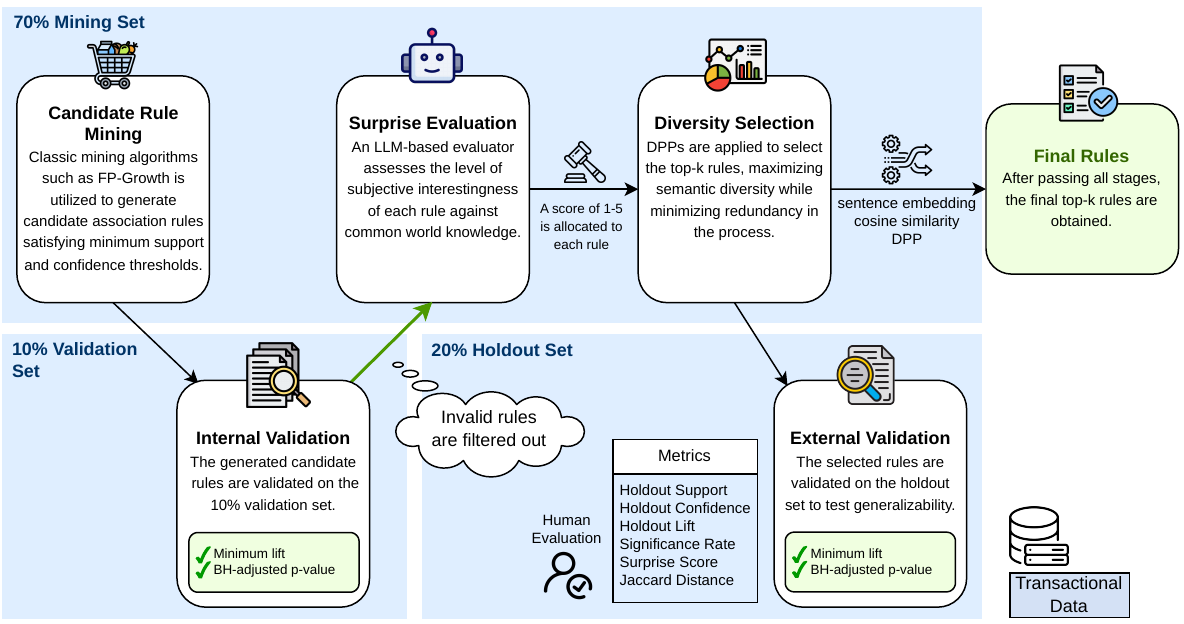}
    \caption{Overview of DiSARM. Transaction data is split into mining, validation, and holdout sets. 
    Candidate rules are extracted and analyzed to produce a final subset of statistically sound rules that are both surprising and diverse.}
    \label{fig:pipeline}
\end{figure*}

Recent advances in Large Language Models (LLMs) offer opportunities to evaluate the subjective interestingness of generated patterns in ADE. In contrast to classical mining algorithms, LLMs can utilize existing world knowledge to reason whether an association is expected or surprising. 
Recent studies have explored LLM-based automated data exploration through agentic approaches~\cite{insightpilot, quis, perez2025}, while others incorporate external domain knowledge into hypothesis generation~\cite{liu2025}. 

Nevertheless, directly applying LLMs in data exploration systems presents some key limitations.  \textbf{(1) Scalability.} A dataset may contain millions of records of raw data, which can overwhelm the LLM's context window. Relying on sampling methods risks overlooking infrequent but potentially significant patterns that are only observable when analyzing the entire dataset. \textbf{(2) Reliability. }Data exploration systems relying on LLMs may contain hallucinated information and logical inconsistencies~\cite{yousuf2026}, while reliability and repeatability are not guaranteed~\cite{labedzki2026}. These limitations suggest that LLMs exhibit better performance when evaluating a set of carefully generated patterns instead of searching for raw associations from scratch.

Extending previous ARM research, we propose DiSARM, a hybrid framework that leverages the world knowledge of LLMs to assess the subjective interestingness of association rules, while maintaining statistical reliability and promoting diversity among the selected rules.
As illustrated in Figure~\ref{fig:pipeline}, classical mining techniques are first applied to the dataset to efficiently extract association rules supported by the dataset. Internal validation is then applied to filter out unreliable rules and enhance generalizability. For the remaining rules, an LLM-based evaluator measures subjective interestingness, leveraging its existing world knowledge to assess the association holistically. Finally, candidate rules are presented as vector embeddings to identify semantic similarities and reduce redundancies, and DPP is applied to select a compact and mutually diverse subset of top-$k$ rules. 

Across three transactional datasets, experimental results demonstrate that DiSARM achieves substantially greater surprise and diversity than classical ARM baselines while retaining stronger statistical validity than LLM baselines. 
Ablation studies further highlight the complementary contributions of each component, demonstrating the framework's ability to deliver statistically reliable, surprising, and diverse patterns for ADE.

\section{Related Work}
Early studies on ARM focused on discovering association rules efficiently and selecting measures to improve statistical validity in the mining data. Further research has focused on more subjective measures of association rules. More recently, LLMs provide opportunities to incorporate external knowledge and discover more subjectively interesting insights.

\subsection{Mining for Interesting Association Rules}
Due to the vast number of rules generated by ARM methods, many studies have explored methods for identifying interesting association rules. One direction focuses on objective interestingness, which measures statistical strengths of relationships among attributes in a dataset~\cite{tan2004}.
Another direction focuses on subjective interestingness, such as eliciting user-defined characteristics of interesting rules~\cite{klemettinen, liu2000}, and incorporating user preferences into the selection of interestingness measures~\cite{lenca2008}. Some studies apply ontology-based techniques and incorporate large-scale knowledge bases to discover novel patterns~\cite{abhilash2023, issahaku2024}, while algorithms have also been developed to extract rules for specific applications~\cite{cohen2001, csaldi2023}.

\subsection{LLM-based EDA Systems}
Autonomous systems are increasingly being developed to conduct exploratory data analysis with limited human guidance. Some studies explore applying LLM-based systems to data science workflows, which convert user requests into executable code~\cite{datainterpreter,taskweaver,lambda}. 
Other works take on a text-to-SQL approach~\cite{mtteql,cheng2023}.
More recent works go a step further, producing natural and coherent human-readable insights~\cite{hypogenic,perez2025}. InsightPilot~\cite{insightpilot} is an LLM-based automated data exploration system, in which the insight engine provides accurate insights to avoid hallucination, while the LLM selects appropriate actions and drives the exploration process. QUIS~\cite{quis} introduces a fully automated EDA framework that consists of two stages: insight generation and question generation. Other studies combine LLMs with data-driven agents to review existing literature and generate novel hypotheses~\cite{liu2025}.

\section{Methodology}
\subsection{Overview of DiSARM}
This work aims to discover association rules that are not only statistically supported by the data but also surprising and diverse. To this end, we propose DiSARM, a novel framework consisting of four main stages. First, candidate association rules are extracted by mining algorithms with minimum support and confidence thresholds. Next, internal validation is conducted to filter out rules that are not statistically significant. An LLM judge evaluates the remaining rules to assess their levels of surprise. Then, the diversity selection stage selects the top-$k$ rules that balance surprise and diversity. During evaluation, the final selected rules are further validated on a holdout set of transactions. 

\subsection{Candidate Rule Mining}

We use classical mining algorithms to reliably generate candidate association rules. In association mining, the problem can be formally stated as follows: Let $I = \{i_1, \, i_2, \, \dots, \, i_m\}$ be a set of items. Let $T$ be a database containing a set of transactions, where each transaction $t$ is a set of items such that $t \subseteq I$. An association rule takes the form $X \rightarrow Y$, where $X \subset I$, $Y \subset I$, and $X \cap Y = \emptyset$. 

Numerous algorithms have been proposed that apply user-specified minimum support and minimum confidence thresholds. The rule $X \rightarrow Y$ holds in the transaction set $T$ with \textit{confidence} $c$ if $c\%$ of transactions in $T$ that support $X$ also support $Y$. The rule has \textit{support} $s$ in $T$ if $s\%$ of the transactions in $T$ contain $X \cup Y$.
Although Apriori~\cite{apriori} is among the most widely implemented algorithms, we adopt FP-Growth because of its efficiency on large transactional datasets. FP-Growth uses a divide-and-conquer method to reduce the search space by recursively constructing conditional pattern bases, substantially reducing search costs~\cite{fpgrowth}. 
These algorithms extract candidate association rules between items in a large database of transactions, avoiding the risk exceeding an LLM's context window. 

\subsection{Internal Validation} 
We split each dataset into a $70\%$ mining set, a $10\%$ internal validation set, and a $20\%$ holdout set. The mining and internal validation sets are available to the framework, whereas the holdout set remains unseen until the final evaluation.

Even though classical ARM algorithms generate rules that satisfy minimum support and confidence thresholds on the mining data, they do not guaranty generalizability to unseen data. In our initial experiments, especially for smaller datasets, a large portion of the top $k$ rules selected through surprise evaluation and diversity selection do not pass statistical validation on the holdout set. To mitigate this, we introduce an internal validation stage before surprise and diversity selection.

To assess generalizability, we compute lift on the validation set for each selected rule "IF [A] THEN [B]". A validation lift value greater than $1$ indicates a positive association, and the occurrence of the antecedent increases the likelihood of the consequent in the unseen data. 

For each rule, we test the null hypothesis $H_0: A \perp B$ against the one-sided alternative $H_1: A \text{ and } B \text{ are positively associated}$. We construct a contingency table (Table~\ref{tab:fisher}) describing the presence or absence of the antecedent (A) and consequent (B) across transactions in the validation set. A one-sided Fisher's exact test with the alternative hypothesis that A and B are positively associated is applied.

\begin{table}[t]
\centering
\small
\caption{Contingency Table for Fisher's Exact Test}
\begin{tabular}{c|c|l}
 & B ($n_2$) & Without B ($N-n_2$)\\\hline
A ($n_1$) & $k$ & $n_1-k$ \\
Without A ($N-n_1$) & $n_2-k$ & $N-n_1-n_2+k$ \\
\end{tabular}
\label{tab:fisher}
\end{table}

We compute the $p$-value using a one-sided Fisher's exact test, summing the hypergeometric probabilities for co-occurrences greater than or equal to the observed count $k$:
\begin{align}
p(R) = \sum_{x=k}^{\min(n_1, n_2)} \frac{\begin{pmatrix} n_1 \\ x \end{pmatrix} \begin{pmatrix} N - n_1 \\ n_2 - x \end{pmatrix}}{\begin{pmatrix} N \\ n_2 \end{pmatrix}}.
\end{align}

Additionally, because multiple rules are evaluated for each method, testing every rule independently at \(p\leq0.05\) would increase the expected number of false positives (Type I errors). To counter this effect, we control the Benjamini-Hochberg false discovery rate (FDR), defined as the expected proportion of false rejections among all rejected hypotheses~\cite{fdr}. For each method, let $m$ denote the number of candidate rules with sorted $p$-values: $p_{(1)}\leq p_{(2)}\leq\cdots\leq p_{(m)}.$ At $\alpha=0.05$ , we find $k=\max
\left\{
i:
p_{(i)}
\leq
\frac{i}{m}\alpha
\right\},$ We then reject the null hypothesis for $p_{(1)},\ldots,p_{(k)}$ and achieve the Benjamini-Hochberg adjusted $p$-values. 

We formally define a rule as statistically validated if it satisfies both of the following criteria on the validation data:
\begin{align*}
q \leq 0.05
\quad\text{and}\quad
\operatorname{lift}_{\mathrm{validation}} > 1,
\end{align*}
where $q$ denotes the Benjamini-Hochberg adjusted p-value. The first condition requires that the association remain statistically significant after correcting for multiple comparisons, while the second requires that the direction of the association remains positive on the unseen data. Only rules that satisfy both conditions are preserved.

\subsection{Surprise Evaluation} 
Association mining algorithms produce a large number of candidate rules that satisfy objective interestingness criteria. However, these measures do not capture whether a rule is unexpected given common world knowledge. We define surprise as a form of subjective interestingness, measured by how difficult an association would have been to anticipate based on ordinary knowledge of consumer behavior. A statistically strong association may still appear obvious to a human analyst, such as the association "customers who purchase cereal are inclined to also purchase milk", and is therefore unlikely to be considered surprising.

To capture subjective interestingness, we deploy an LLM-based evaluator to allocate a surprise score of  $1$ to $5$ for each candidate association rule, depending on how difficult the rule is to anticipate given existing world knowledge. The evaluator is prompted to explain whether a potential relationship underlying the hypothesis can be anticipated and provide a step-by-step explanation to support the score given. This explanation encourages the model to explicitly reason whether the relationship between the antecedent and consequent is already predictable from common knowledge.

\captionof{prompt}{Prompt of Surprise Evaluation.}
\label{prompt:surprise}
\begin{llmprompt}
You are an expert Data Science LLM Judge evaluating association rules derived from transactional data. Your task is to rate how surprising the given association rule is. Surprise is defined as how difficult the association would have been to anticipate before seeing the data, based on ordinary knowledge of consumer behavior. Here is the rule: \{rule\_string\}. Please rate the association from 1 (The association is immediately expected from common knowledge) to 5 (Highly unexpected and genuinely difficult to anticipate from ordinary knowledge alone), using the format:\\
\{{"surprise\_assessment": "<one brief sentence of explanation>","surprise\_score": <integer: 1 to 5>\}}

\end{llmprompt}

\subsection{Diversity Selection} 
Selecting rules based on statistical strength or surprise can still result in a set of semantically similar or redundant rules, as association rules may differ in phrasing but express similar relationships. Therefore, we introduce a diversity selection stage based on Determinantal Point Processes (DPPs) to select the final top-$k$ rules.

First, each candidate rule is converted into a natural-language sentence and embedded using the pretrained Sentence-BERT model \texttt{all-MiniLM-L6-v2}. Semantic similarity between rules $i$ and $j$ is quantified using cosine similarity $S_{ij}$:
\begin{equation}
S_{ij} = \frac{\mathbf{e}_i^\top \mathbf{e}_j}{\|\mathbf{e}_i\|_2 \|\mathbf{e}_j\|_2},
\end{equation}
where $\mathbf{e}_i$ and $\mathbf{e}_j$ denote the corresponding rule embeddings.

To construct the DPP kernel, candidate rules are ranked according to their surprise scores obtained from the previous stage, with lift and confidence used as successive tie-breakers. The quality weight for rule $i$ is calculated as:
\begin{equation}
q_i = \exp\left(\alpha\left(1-\frac{p_i}{n-1}\right)\right),
\end{equation}
where $p_i$ denotes the zero-based quality rank of rule $i$, $n$ is the number of candidate rules, and $\alpha$ is set to $0.5$. Following the quality-weighted DPP formulation~\cite{kulesza2012}, the kernel is defined as:
\begin{equation}
L_{ij} = q_i S_{ij} q_j.
\end{equation}
We then seek to approximately solve the following optimization problem:
\begin{equation}
Y^* \approx \arg\max_{Y \subseteq \mathcal{R}, |Y|=k}
\det(L_Y),
\end{equation}
where $\mathcal{R}$ denotes the candidate rule set and $L_Y$ is the principal submatrix of $L$ corresponding to subset $Y$. We employ fast greedy maximum a posteriori (MAP) inference~\cite{chen2018} to approximate the optimal subset. This allows the framework to select a set of mutually dissimilar rules while prioritizing those with high subjective surprise.

\section{Experiments}
To evaluate the performance of DiSARM, we conduct experiments using GPT-5.1 and implement various classical ARM and LLM baselines, selecting the top-$k$ rules with $k=10$. We investigate the following key questions through experiments:

\begin{itemize}
    \item \textbf{RQ1}: What advantages does DiSARM offer over classical statistics and LLM baselines?
    \item \textbf{RQ2}: To what extent can DiSARM identify rules that are more surprising and diverse while maintaining statistical validity?
    \item \textbf{RQ3}: How effective is the LLM-based surprise evaluation in selecting rules that contain information considered unexpected with respect to general common sense?
    \item \textbf{RQ4}: How does diversity selection contribute to reducing redundancy of the selected rules?
\end{itemize}

\subsection{Datasets}
We conduct experiments on three transactional datasets: Instacart ($3,346,083$ transactions) ~\cite{instacart}, Retail Market Basket Transactions Dataset ($7,501$ transactions) \cite{retail}, and Groceries Market Basket Dataset ($9,835$ transactions) \cite{groceries}.

\subsection{Evaluation Metrics and Holdout Validation}
We evaluate the discovered association rules using a holdout dataset, assessing their statistical soundness, surprise, and diversity.\\
\textbf{Soundness.} 
Although classical ARM algorithms generate rules that satisfy minimum support and confidence thresholds on the mining data, these criteria do not guarantee generalizability to unseen data. To assess whether the discovered associations are not simply LLM hallucinations or local noise, we report support, confidence, and lift on the holdout set. For the association rule "IF [A] THEN [B]", the metrics are defined as:
\begin{align*}
\mathrm{Support}(A\Rightarrow B) =P(A\cup B),\\
\mathrm{Confidence}(A\Rightarrow B) =\frac{P(A\cup B)}{P(A)},\\
\mathrm{Lift}(A\Rightarrow B) =\frac{P(A\cup B)}{P(A)P(B)}.      
\end{align*}
Furthermore, we report the significance rate, defined as the percentage of the final selected rules that are statistically validated on the unseen holdout set, satisfying the following criteria:
\begin{align*}
q \leq 0.05
\quad\text{and}\quad
\operatorname{lift}_{\mathrm{holdout}} > 1.
\end{align*}

\textbf{Surprise.} 
We use GPT-5.1 as the LLM-based surprise scorer to identify candidate rules that are likely to contain subjectively interesting information. For evaluation, we use Gemini-3.1-Pro as an independent LLM judge to assess the level of surprise in the final selected rules. We select Gemini-3.1-Pro, developed by Google, to provide an independent evaluation using a model from a different provider. This separation reduces dependence on a single model family and allows for an external assessment of surprise.

\textbf{Diversity.}
We evaluate the diversity of the final selected rule set using Jaccard distance. Each association rule is represented by the union of the items appearing in its antecedent and consequent.
For two rules $r_i$ and $r_j$, their Jaccard distance is defined as:
\begin{equation}
d_J(r_i,r_j)=1-\frac{|r_i\cap r_j|}{|r_i\cup r_j|},
\end{equation}
which measures the overlap between the two sets. We report the mean pairwise Jaccard distance across all pairs of selected rules.
A higher value indicates rules share fewer common items and are less redundant. While DPP in the diversity selection stage utilizes vector embeddings and cosine similarity, Jaccard distance provides an item-level assessment of diversity.

\subsection{Baselines}
\textbf{Classical ARM baselines.}
While our approach aims to discover rules that are subjectively interesting given prior world knowledge, many existing metrics aim to evaluate objective interestingness, such as a rule's structure, predictive performance, and statistical significance.
Although classical ARM algorithms prune rules using support and confidence thresholds, other metrics capture different and sometimes contrasting aspects of interestingness~\cite{interestingness}.

For the classical ARM baselines, we select rules based on lift, Piatetsky-Shapiro's (PS) measure, or J-Measure (J):
\begin{align*}
\mathrm{Lift}(A\Rightarrow B)
&=\frac{P(A\cap B)}{P(A)P(B)},\\
\mathrm{PS}(A\Rightarrow B)
&=P(A\cap B)-P(A)P(B),\\
\mathrm{J}(A\Rightarrow B)
&=P(A)\Big[
P(B\mid A)\log_2\tfrac{P(B\mid A)}{P(B)}\\
&\qquad+
P(\neg B\mid A)\log_2\tfrac{P(\neg B\mid A)}{P(\neg B)}
\Big].
\end{align*}
Lift measures the strength of an association relative to the expected co-occurrence under independence. PS measures the difference between the observed and the expected co-occurrence probabilities under independence, and has been found to outperform lift when noisy products are filtered out~\cite{kirchgessner}. J-Measure is an information-theoretic measure that quantifies the information the antecedent provides about the consequent~\cite{blanchard}.
After generating candidate rules using FP-Growth, we rank and select rules according to each of these established objective interestingness metrics.

\textbf{LLM baselines.} We adapt the Serial framework proposed by~\cite{perez2025} to our task, which serializes a subset of selected transactions into an HTML table and passes the table to the LLM due to context window limitations~\cite{jiang2024}. In our experiments, we implement two adaptations: (1) \textbf{Serial Standard}, which prompts the model to output 10 association rules directly from the selected subset, and (2) \textbf{Serial Surprise}, which prompts the LLM to focus on rules that are statistically significant but whose underlying relationships are not immediately apparent. In other words, Serial Surprise explicitly elicits rules that contain an element of surprise.

\subsection{Results}
In this section, we compare DiSARM with classical ARM and LLM baselines to evaluate its ability to balance statistical validity, surprise, and diversity. For the data split, we use five random seeds: $42, 943, 3136, 4799$, and $8315$, and report the mean and standard deviation values of each metric over the five runs. Results are detailed in Tables~\ref{tab:metrics_instacart}-\ref{tab:metrics_groceries}.

\begin{table*}[t]
    \centering
    \footnotesize
    \setlength{\tabcolsep}{2pt}
    \begin{tabular*}{\textwidth}{@{\extracolsep{\fill}}l|cccccc@{}}
        \toprule
        Method & Support & Confidence & Lift & Significance Rate & Surprise & Jaccard Distance \\
        \midrule
        Classical Lift & $0.001 \pm 0.000$ & $0.33 \pm 0.01$ & $\bf{73.05 \pm 1.21}$ & $\bf{1.00 \pm 0.00}$ & $1.00 \pm 0.00$ & $0.74 \pm 0.00$ \\
        Classical PS & $\bf{0.014 \pm 0.000}$ & $0.28 \pm 0.01$ & $2.52 \pm 0.16$ & $\bf{1.00 \pm 0.00}$ & $1.00 \pm 0.00$ & $0.89 \pm 0.01$ \\
        Classical J-Measure & $0.005 \pm 0.000$ & $\bf{0.34 \pm 0.01}$ & $42.37 \pm 2.57$ & $\bf{1.00 \pm 0.00}$ & $1.00 \pm 0.00$ & $0.88 \pm 0.02$ \\
        \midrule
        Serial Standard & $0.007 \pm 0.001$ & $0.11 \pm 0.01$ & $2.47 \pm 0.44$ & $0.98 \pm 0.05$ & $1.02 \pm 0.04$ & $0.93 \pm 0.03$ \\
        Serial Surprise & $0.000 \pm 0.000$ & $0.03 \pm 0.03$ & $4.58 \pm 5.26$ & $0.44 \pm 0.29$ & $\bf{3.38 \pm 0.82}$ & $\bf{0.99 \pm 0.01}$ \\
        \midrule
        \bf{DiSARM} & $0.002 \pm 0.000$ & $0.26 \pm 0.00$ & $3.97 \pm 0.03$ & $\bf{1.00 \pm 0.00}$ & $2.00 \pm 0.00$ & $0.93 \pm 0.00$ \\
        \bottomrule
    \end{tabular*}
    \caption{Comparison of DiSARM against classical ARM and LLM baselines on Instacart.}
    \label{tab:metrics_instacart}
\end{table*}

\begin{table*}[t]
    \centering
    \footnotesize
    \setlength{\tabcolsep}{2pt}
    \begin{tabular*}{\textwidth}{@{\extracolsep{\fill}}l|cccccc@{}}
        \toprule
        Method & Support & Confidence & Lift & Significance Rate & Surprise & Jaccard Distance\\
        \midrule
        Classical Lift & $0.019 \pm 0.001$ & $0.27 \pm 0.01$ & $\bf{2.31 \pm 0.11}$ & $0.92 \pm 0.08$ & $1.40 \pm 0.12$ & $0.87 \pm 0.01$ \\
        Classical PS & $\bf{0.047 \pm 0.001}$ & $0.31 \pm 0.01$ & $1.63 \pm 0.05$ & $0.96 \pm 0.09$ & $1.88 \pm 0.08$ & $0.71 \pm 0.03$ \\
        Classical J-Measure & $0.029 \pm 0.003$ & $\bf{0.32 \pm 0.01}$ & $2.11 \pm 0.09$ & $\bf{0.98 \pm 0.05}$ & $1.72 \pm 0.11$ & $0.87 \pm 0.03$ \\
        \midrule
        Serial Standard & $0.031 \pm 0.002$ & $0.22 \pm 0.03$ & $1.58 \pm 0.14$ & $0.62 \pm 0.18$ & $1.60 \pm 0.35$ & $0.88 \pm 0.04$ \\
        Serial Surprise & $0.002 \pm 0.001$ & $0.19 \pm 0.02$ & $1.98 \pm 0.64$ & $0.10 \pm 0.14$ & $\bf{4.28 \pm 0.36}$ & $\bf{0.95 \pm 0.07}$ \\
        \midrule
        \bf{DiSARM} & $0.022 \pm 0.001$ & $0.29 \pm 0.02$ & $1.65 \pm 0.08$ & $0.74 \pm 0.11$ & $2.78 \pm 0.36$ & $\bf{0.95 \pm 0.01}$ \\
        \bottomrule
    \end{tabular*}
    \caption{Comparison of DiSARM against classical ARM and LLM baselines on Retail.}
    \label{tab:metrics_retail}
\end{table*}

\begin{table*}[t]
    \centering
    \footnotesize
    \setlength{\tabcolsep}{2pt}
    \begin{tabular*}{\textwidth}{@{\extracolsep{\fill}}l|cccccc@{}}
        \toprule
        Method & Support & Confidence & Lift & Significance Rate & Surprise & Jaccard Distance \\
        \midrule
        Classical Lift & $0.020 \pm 0.004$ & $0.30 \pm 0.02$ & $\bf{2.28 \pm 0.05}$ & $0.98 \pm 0.05$ & $1.06 \pm 0.09$ & $0.90 \pm 0.03$ \\
        Classical PS & $\bf{0.054 \pm 0.003}$ & $0.34 \pm 0.01$ & $1.74 \pm 0.05$ & $\bf{1.00 \pm 0.00}$ & $1.12 \pm 0.04$ & $0.69 \pm 0.02$ \\
        Classical J-Measure & $0.040 \pm 0.003$ & $\bf{0.37 \pm 0.01}$ & $2.04 \pm 0.06$ & $\bf{1.00 \pm 0.00}$ & $1.10 \pm 0.00$ & $0.82 \pm 0.02$ \\
        \midrule
        Serial Standard & $0.033 \pm 0.005$ & $0.26 \pm 0.04$ & $1.54 \pm 0.02$ & $0.84 \pm 0.11$ & $1.04 \pm 0.09$ & $0.88 \pm 0.03$ \\
        Serial Surprise & $0.001 \pm 0.000$ & $0.09 \pm 0.05$ & $1.12 \pm 0.38$ & $0.02 \pm 0.05$ & $\bf{4.04 \pm 0.47}$ & $\bf{0.98 \pm 0.01}$ \\
        \midrule
        \bf{DiSARM} & $0.018 \pm 0.002$ & $0.28 \pm 0.02$ & $1.81 \pm 0.02$ & $0.84 \pm 0.06$ & $1.64 \pm 0.15$ & $0.97 \pm 0.01$ \\
        \bottomrule
    \end{tabular*}
    \caption{Comparison of DiSARM against classical ARM and LLM baselines on Groceries.}
    \label{tab:metrics_groceries}
\end{table*}

\begin{figure*}[t]
    \centering
    \includegraphics[width=\textwidth, trim={7mm 0mm 7mm 0mm}]{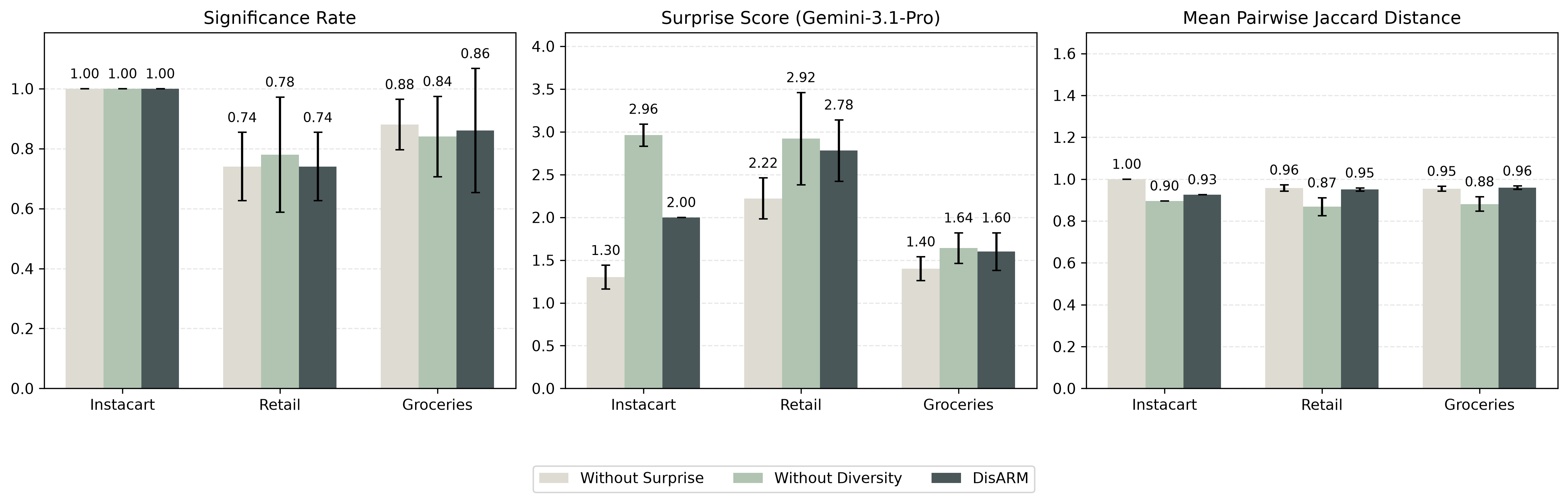}
    \caption{Comparison of significance rate, surprise scores evaluated by Gemini-3.1-Pro, and mean pairwise Jaccard distance across three settings: DiSARM, without surprise evaluation, and without diversity selection.}
    \label{fig:ablation}
\end{figure*}

\textbf{Comparison with Classical ARM Baselines (RQ1-1).} The three classical ARM baselines achieve the strongest statistical validity. Their significance rates do not drop below $0.92$, and the Lift and J-measure baselines achieve a significantly higher holdout lift. 
However, these statistically strong rules are considerably less surprising and diverse than those selected by DiSARM. On Instacart, all three baselines receive a surprise score of $1.00$, compared with $2.00$ for DiSARM. DiSARM also achieves a Jaccard distance of $0.95$ on Retail, higher than the classical ARM baselines, which range from $0.71$ to $0.87$.
These results indicate that objective interestingness measures used by classical ARM baselines are effective at identifying statistically strong rules. However, DiSARM achieves substantially higher surprise and diversity at the cost of some holdout lift and statistical significance on smaller datasets.

\textbf{Comparison with LLM Baselines (RQ1-2).}
The LLM baselines reveal a trade-off between statistical validity and surprise. Serial Standard achieves relatively high statistical validity but low surprise scores. For example, it achieves a significance rate of $0.98$ on Instacart, while its surprise score remains at $1.02$. This suggests that sequential rule generation without explicit surprise selection does not consistently lead to highly unexpected rules.
In contrast, Serial Surprise achieves the highest surprise scores among all compared methods, with $3.38$ on Instacart, $4.28$ on Retail, and $4.04$ on Groceries. It also obtains high Jaccard distances of $0.99$, $0.95$, and $0.98$, respectively. However, its significance rate drops to $0.44$ on Instacart, $0.10$ on Retail, and only $0.02$ on Groceries. These findings underline the difficulty of achieving both subjective surprise and statistical validity through LLM-based association mining alone.

\textbf{Analysis of the Proposed Method (RQ2).} 
The overall results illustrated in Tables~\ref{tab:metrics_instacart}-\ref{tab:metrics_groceries} show that DiSARM achieves a more balanced combination of statistical validity, surprise, and diversity. Compared with the classical ARM baselines, it yields more surprising and diverse rules, whereas compared with Serial Surprise, its selected rules have substantially higher statistical significance. For example, on Instacart, DiSARM achieves a significance rate of $1.00$, a surprise score of $2.00$, and a Jaccard distance of $0.93$. These results indicate that the proposed framework does not maximize any single metric individually. Instead, by combining classical association rule mining with LLM-based surprise evaluation and diversity selection, it identifies rules that are simultaneously statistically supported, subjectively surprising, and mutually diverse.

\subsection{Human Evaluation}
To assess the agreement between Gemini-3.1-Pro and human judgments of surprise, we conduct a human evaluation of the association rules selected by DiSARM and the baseline methods using random seed $42$. Across all methods and datasets, a total of $240$ rules are selected, corresponding to $170$ distinct rules. Each distinct rule is assigned a human surprise score on a scale of 1 to 5, following the evaluation instructions detailed in Appendix~\ref{sec:human}. 

Across all datasets, the evaluation yields a Pearson linear correlation coefficient of $r = 0.72$ ($p = 4.89 \times 10^{-28}$) and a Spearman’s rank correlation coefficient of $\rho = 0.67$ ($p = 1.14 \times 10^{-23}$). Positive correlations are also observed across all datasets. These results indicate consistent positive alignment between human judgments and surprise scores assigned by Gemini-3.1-Pro, which supports the use of our LLM-as-a-judge framework as an estimated measure of subjective interestingness.

\subsection{Ablation Study}

In this section, we conduct ablation experiments by removing either surprise evaluation or diversity selection from the proposed methodology. Results are shown in Figure~\ref{fig:ablation}.

\textbf{Effectiveness of Surprise Evaluation (RQ3).} To evaluate the effectiveness of LLM-based surprise evaluation, we compare DiSARM with one that removes the surprise evaluation stage. Across all three datasets, removing surprise evaluation decreases the surprise score from Gemini-3.1-Pro. In particular, the surprise score decreases from $2.00$ to $1.30$ on Instacart. These results indicate that the LLM-based filtering stage effectively places greater emphasis on rules that are judged to be more surprising and less consistent with everyday expectations.

We observe that removing surprise evaluation generally increases the Jaccard distance, since the remaining diversity selection component directly favors dissimilar rules. Conversely, the significance rate remains comparable to that of the proposed method across the three datasets, indicating that surprise evaluation does not have a clear negative impact on statistical validity.

\textbf{Effectiveness of Diversity Selection (RQ4).}
We next evaluate the role of diversity selection by removing it while retaining surprise evaluation. The surprise score is generally higher with diversity selection removed, but the Jaccard distance decreases consistently across all three datasets. For example, DiSARM achieves a Jaccard distance of $0.93$ on Instacart, but removing diversity selection reduces it to $0.667$.

Although optimizing for diversity compromises the surprise score, DiSARM maintains relatively high surprise while producing a more diverse set of rules. The significance rate remains broadly comparable across the three settings, indicating that this trade-off is achieved without a clear loss in statistical validity.

Overall, the ablation results demonstrate that surprise evaluation and diversity selection are two complementary stages. Combining both components provides a balance between surprise and diversity while maintaining statistical significance on the holdout data.

\section{Conclusion}
This paper focuses on association rule mining for discovering surprising and diverse top-$k$ insights. We propose a framework that employs classical ARM algorithms to extract statistically supported rules while incorporating LLM-based surprise evaluation and diversity subset selection. By comparing the proposed framework with classical ARM baselines, including Lift, PS, and J-Measure, as well as LLM baselines, we show that DiSARM achieves a balance among statistical validity, subjective surprise, and diversity.

\clearpage

\section{Limitations}
This work introduces a novel framework that extracts and selects association rules using classical ARM algorithms and an LLM-as-a-judge approach, achieving a balance among statistical validity, subjective surprise, and mutual diversity. However, several limitations remain. i) Requiring statistical significance and minimum support and confidence thresholds improves validity, but rarer associations may fail validation and be neglected despite containing potentially surprising information. ii) As surprise is an inherently subjective and culturally dependent concept, its evaluation cannot be treated as ground truth. Even with human evaluation, surprise scores should be viewed as a representation of subjective interestingness, and conclusions should be interpreted as relative rather than absolute.

\section{AI Assistance Disclosure}
AI assistants were utilized to support in code development and manuscript editing.
The authors thoroughly reviewed and verified all generated content.

\bibliography{custom}

\appendix

\section{Implementation Details}
\label{implementation}
We implement FP-Growth using the \texttt{mlxtend} library (version 0.25.0). The minimum support thresholds are set to $0.001$ for Instacart and $0.01$ for Retail and Groceries, while the minimum confidence threshold is set to $0.3$ across all datasets.

GPT-5.1 and Gemini-3.1-Pro are accessed through OpenRouter. As both are proprietary models, their exact parameter counts are not publicly disclosed. Rule embeddings are generated using \texttt{all-MiniLM-L6-v2}, which contains approximately 22.7 million parameters, through the \texttt{sentence-transformers} library (version 5.6.0).

\section{Human Evaluation Details}
\label{sec:human}
To evaluate subjective interestingness, we manually score the level of surprise for the final selected association rules. The following is the instructions and rubric used for scoring:

\captionof{instruction}{Instructions of Human Evaluation for Surprise.}
\label{instruction:human}
\begin{llmprompt}
Your task is to rate the level of surprise of a given association rule.\\
Surprise Definition:
How difficult would this association have been to anticipate before seeing the data?\\
Surprise Scale:\\
1 (Obvious): Immediately expected from common knowledge or a direct conventional relationship between the items.\\
2 (Expected): The specific pairing is not inevitable, but the items have a clear shared category, use, consumption context, or shopping relationship that makes the association fairly predictable.\\
3 (Moderately Surprising): The association would not be an immediate prediction, but a reasonable connection between the items is readily apparent.\\
4 (Surprising): The association would normally be difficult to predict beforehand, although a plausible connection can be found upon closer consideration.\\
5 (Highly Surprising): The items have little or no obvious behavioral, functional, or consumption-related relationship, making the association genuinely difficult to anticipate without seeing the data.
\end{llmprompt}

\section{Prompts for LLM Baselines}
\label{sec:llm_baseline}
Prompt~\ref{prompt:serial_standard} and Prompt~\ref{prompt:serial_surprise} present the detailed prompts for Serial Standard and Serial Surprise, respectively. Bold text highlights the modifications made for the Serial Surprise prompt.

\captionof{prompt}{Prompt of Serial Standard.}
\label{prompt:serial_standard}
\begin{llmprompt}
You are an expert data analyst performing exploratory analysis of market-basket transaction data.
You will receive an HTML table containing a random subset of market baskets. Each row represents one transaction, and products purchased together are separated by the "|" character.
Your task is to directly inspect the HTML table and generate exactly \{k\} product association rules.
Each rule must have the form: IF [product A] THEN [product B].\\

Return valid JSON using this structure:\\
\{
  "rules": [
    \{
      "antecedent": "product A",
      "consequent": "product B",
      "rule\_string": "IF [product A] THEN [product B]",
      "reasoning": "<one brief sentence describing why this association appears plausible from the supplied transactions>"
    \}
  ]
\}
\end{llmprompt}

\captionof{prompt}{Prompt of Serial Surprise.}
\label{prompt:serial_surprise}
\begin{llmprompt}
You are an expert data analyst performing exploratory analysis of market-basket transaction data.
You will receive an HTML table containing a random subset of market baskets. Each row represents one transaction, and products purchased together are separated by the "|" character.
Your task is to directly inspect the HTML table and generate exactly {k} unexpected association rules. \textbf{Focus on pairs that co-occur but lack an obvious or commonly expected relationship, but should also be statistically significant given the subset of market baskets.}
Each rule must have the form: IF [product A] THEN [product B].\\

Return valid JSON using this structure:\\
\{
  "rules": [
    \{
      "antecedent": "product A",
      "consequent": "product B",
      "rule\_string": "IF [product A] THEN [product B]",
      "reasoning": "\textbf{<one brief sentence stating why this association is unexpected yet present in the data>}"
    \}
  ]
\}
\end{llmprompt}

\end{document}